# Phenotyping calcification in vascular tissues using artificial intelligence

Mehdi Ramezanpour [a], Anne M. Robertson [a*], Yasutaka Tobe [a], Xiaowei Jia [b], Juan R. Cebral [c]

[a] Department of Mechanical Engineering and Materials Science, University of Pittsburgh, PA, USA
[b] Department of Computer Science, University of Pittsburgh, PA, USA
[c] Department of Mechanical Engineering, George Mason University, Fairfax, Virginia, USA

* Address correspondence to Anne M. Robertson, Department of Mechanical Engineering and Materials Science, University of Pittsburgh, Pittsburgh, PA, USA.
**Email:**  rbertson@pitt.edu

## Abstract

Vascular calcification is implicated as an important factor in major adverse cardiovascular events (MACE), including heart attack and stroke.  A controversy remains over how to integrate the diverse forms of vascular calcification into clinical risk assessment tools.  Even the commonly used calcium score for coronary arteries, which assumes risk scales positively with total calcification, has important inconsistencies.  Fundamental studies are needed to determine how risk is influenced by the diverse calcification phenotypes. However, studies of these kinds are hindered by the lack of high-throughput, objective, and non-destructive tools for classifying calcification in imaging data sets. Here, we introduce a new classification system for phenotyping calcification along with a semi-automated, non-destructive pipeline that can distinguish these phenotypes in even atherosclerotic tissues.  The pipeline includes a deep-learning-based framework for segmenting lipid pools in noisy µ-CT images and an unsupervised clustering framework for categorizing calcification based on size, clustering, and topology. This approach is illustrated for five vascular specimens, providing phenotyping for thousands of calcification particles across as many as 3200 images in less than seven hours.  Average Dice Similarity Coefficients of 0.96 and 0.87 could be achieved for tissue and lipid pool, respectively, with training and validation needed on only 13 images despite the high heterogeneity in these tissues.  By introducing an efficient and comprehensive approach to phenotyping calcification, this work enables large-scale studies to identify a more reliable indicator of the risk of cardiovascular events, a leading cause of global mortality and morbidity.

**Significance Statement**

Risk assessment tools for major adverse cardiovascular events (MACE) have recognized limitations, including simplistic accounting for calcification. Using high-resolution intravascular imaging, it is becoming possible to distinguish diverse calcification forms in vivo.  However, fundamental studies of the distinct mechanistic roles of these phenotypes are needed before we can leverage this wealth of medical data to improve both risk assessment and treatment. Such studies require high-throughput, objective, and non-destructive tools for phenotyping calcification. Here, we introduce a framework that meets these criteria along with a new classification system for phenotyping calcification. This semi-automated, non-destructive pipeline uses deep-learning-based segmentation and unsupervised clustering algorithms to provide phenotyping for thousands of calcification particles across more than 3200 µ-CT images in less than seven hours.

**1- Introduction**

Arterial calcifications are linked to major adverse cardiovascular events (MACE) that are categorized as the leading causes of mortality and morbidity worldwide (1-4). For instance, in silico studies have shown calcifications in fibrous caps within atherosclerotic plaques can impact cap vulnerability (5, 6). The rupture of these fibrous caps releases the plaque contents into the bloodstream, resulting in the formation of thrombi, which can obstruct blood flow, leading to serious health issues, including ischemic stroke and myocardial infarction. Calcifications also play a pivotal role in the failure of the aortic valve, culminating in left ventricle hypertrophy and, ultimately, heart failure (1, 7).

Arterial calcifications exhibit phenotypes that can significantly modulate the failure of diseased tissues (3, 4, 6, 8-10). For example, some computational studies reported macrocalcifications can stabilize plaques in coronary arteries (8, 11, 12). On the other hand, a recent in silico simulation shows macrocalcifications with jagged edges can create regions of stress concentrations in cerebral aneurysms, promoting tissue failure. Therefore, macrocalcifications can be either protective or detrimental, depending on their topology. Regarding microcalcifications, in silico studies suggest they can induce stress concentrations in the surrounding matrix (13, 14). Moreover, an experimental study found an association between high densities of microcalcifications and low levels of collagen fibers. Since collagen fibers are the main load-bearing components of soft tissues under physiological conditions, this suggests a second mechanism by which microcalcifications can lower the load-bearing capacity of these tissues (15). An additional calcification phenotype arises from the milieu in which calcifications are embedded. A recent experimental analysis shows that ruptured aneurysms with calcification always lacked protective lipid pools (only nonatherosclerotic) (3). These calcification phenotypes can act in a coupled manner to impact the failure process differently than a simple superposition of individual phenotypes. For example, computational studies have reported failure of calcified vascular tissue is strongly influenced by the environment neighboring the calcification. When calcifications are embedded in lipid pools (atherosclerotic calcifications), the tissue failure process was found to be more gradual, potentially enabling tissue repair, in contrast to the abrupt fiber failure found in tissues with non-atherosclerotic calcification (16).

Accounting for the influence of these diverse calcification phenotypes in risk assessment tools, for example, for MACE, can potentially overcome the current limitations of these tools.  For example, the commonly used calcium score, which assumes risk scales positively with total calcification, has important inconsistencies that are the subject of increasing attention (17-20). It is found the calcium score increases when statins are used, despite statins being known to decrease the risk for cardiovascular disease (21, 22). In this regard, including calcification density in the calcium score could significantly improve the accuracy of risk stratification. This underscores the potential to improve the diagnostic values of such risk assessment tools by integrating additional calcification phenotypes (23).

To incorporate these diverse forms of calcification into risk assessment tools for adverse cardiovascular events, an initial step should involve determining the impacts of various calcification phenotypes on tissue strength. Computational analyses, while valuable for assessing many of the mechanical aspects of vascular tissues, rely on mathematical models and simplifying assumptions about the underlying physical processes. Therefore, it becomes crucial to complement these computational approaches with experimental investigations aimed at validating the extent to which these models accurately reflect the actual mechanical behavior of calcified vascular tissues. These experimental analyses can be conducted on dissected samples of calcified vascular tissues, allowing for an in-depth exploration of the intricate interactions between realistic combinations of calcification phenotypes and other structural components within the tissue, thereby enhancing our overall understanding. If conducted on a sufficiently large sample set, such analysis could unveil correlations between calcification phenotypes and tissue vulnerability, and ultimately provide the required information to improve the diagnostic value of risk assessment tools.

Novel pharmacological treatments are being developed to reduce calcifications in vascular tissues (24). While these treatments have tremendous clinical potential, extensive scientific studies are needed to understand how the vascular calcifications change in treated tissues. Such studies traditionally were carried out using histological analysis of serial sections (25), a destructive and highly laborious approach for heterogeneous tissues, resulting in substantial sectioning artifacts. In some cases, high-resolution µ-CT imaging is used, though a typical dissected arterial sample can comprise thousands of slices, making the current manual approaches unfeasible for large-scale studies. Therefore, there is a pressing need to implement automated high-throughput approaches to process such large datasets.

In this work, we present a novel approach that addresses these needs by employing advanced machine learning algorithms to high-resolution µ-CT images in a semi-automatic pipeline. We introduce a new categorization system for calcification phenotypes based on size (macro/micro), spatial distribution of microcalcifications (clustered/isolated), topology of macrocalcifications (sparse/dense) and co-localization with lipid pools (atherosclerotic/non-atherosclerotic). In the context of macrocalcifications, the term "sparse" pertains to areas that exhibit thin or porous characteristics, whereas the term "dense" applies to the remaining regions. Sparse regions are distinguished by their thin or porous nature, whereas the remaining portions are categorized as dense. This subcategorization of macrocalcifications draws inspiration from the model presented by Hutcheson, *et al.* (15) to explain the genesis of macrocalcifications. This model, which is derived from in-vivo experimental analysis, outlines a stepwise progression for the formation of macrocalcifications from microcalcifications.

To classify calcification across specimens, we start with high-resolution µ-CT to obtain 3D data for tissue, lipid pools and calcification particles. Then, we apply a semi-automatic deep learning-based segmentation algorithm to thousands of µ-CT images to identify tissue and lipid pool regions. Despite naturally noisy µ-CT datasets, our algorithm achieved a Dice similarity coefficient >0.80 compared to human experts. Finally, we classify calcifications based on size, microcalcification distribution, macrocalcification topology, and type via an unsupervised clustering algorithm resulting in eight distinct phenotypes (Fig. 1).

We use two artificial intelligence (AI) algorithms to create a fast and reliable tool for phenotyping vascular calcifications in the high-resolution µ-CT datasets. We then apply this tool to five vascular samples. Processing a specimen containing more than 5000 calcifications was accomplished in under seven hours. We then demonstrate an additional application of the machine learning tool to the growing area of biomechanics of collagen fibers in soft tissues (26), finding a correlation between clustered non-atherosclerotic microcalcifications and local degradation in the structural integrity of collagen fibers. This result is consistent with findings that consider collagen fibers as a structural framework for facilitating calcification formation (15). In areas where collagen fibers are less concentrated and do not impose significant restrictions, calcifications exhibit a greater propensity to aggregate densely and fuse together, ultimately forming macrocalcifications (15).

Using high-resolution intravascular imaging tools, it is becoming possible to distinguish diverse forms of calcification in vivo (27). However, fundamental studies of the distinct mechanistic roles of these phenotypes are needed before we can leverage this wealth of medical data to improve both risk assessment and treatment. The machine learning pipeline introduced here makes it possible, for the first time, to efficiently and reliably phenotype calcifications across entire intact arterial specimens. These tools can be used in fundamental investigations of the role of calcification in soft tissue failure as well as to identify correlations between calcification phenotype and patient characteristics. Such information is vital if clinicians are to be provided with a more sophisticated scoring tool that goes beyond total calcification volume by differentiating between destabilizing and protective calcification.

**2- Results**

**2.1 Deep learning framework for segmentation of sample and lipid pool**

We introduce a novel hybrid neural network framework designed to accurately and semi-automatically segment both the sample and embedded lipid pools from the $\mu$-CT datasets for vascular tissues (Fig. 2). This approach combines the capabilities of a deep learning model UNet designed specifically for segmenting biomedical images (28) with two sequentially connected (neural) networks. The framework aims to efficiently segment diverse vascular specimens with highly heterogeneous composition in a semi-automatic fashion using only a few hand labeled slices from a stack of thousands of images.

Briefly, the proposed framework utilizes transfer learning, employing UNet as a feature extractor to train the initial neural network (sample classifier) responsible for distinguishing foreground (sample domain) from the background. Training is performed on a small subset of images consisting of 13 slices with manually annotated markings of the sample and lipid pool boundaries. During the training process, the kernel weights in the convolutional layers of UNet are optimized to assign higher probability values to pixels belonging to sample regions compared to other regions. This 2D feature map is then converted to a tall vector and is utilized as input feature along with other features, including pixel 3D spatial information (inter and intra-slice coordinates) and their grey-scale intensity, to train the sample classifier. The inclusion of pixel 3D spatial information enables the framework to leverage the continuity of the volumetric variations across regions of interest, enabling automatic identification of intra-slice features shared by foreground regions between slices. The resulting segmentation maps are then utilized to extract sample regions from the dark background of the input image, thereby reducing the number of pixels involved in the training process of the second dense-layered neural network classifier (lipid pool classifier), which is responsible for distinguishing lipid and non-lipid pool pixels within the sample region. By filtering out background pixels, manually extracting input features, and considering inter and intra-slice coordinates of pixels, our framework empowers the lipid pool classifier with sufficient predictive power to accurately classify lipid and non-lipid pixels.

**2.2 Performance assessment- Reliable segmentation of thousands of high-resolution µ-CT images by training on only a few hand-labeled images**

The performance of the deep learning framework was evaluated on specimens from three intracranial arteries and two cerebral aneurysms. These five samples (denoted 1-5) were scanned using high-resolution µ-CT with a three-micron spacing with volumes ranging from 4 – 83 mm$^3$, generating a range of stack sizes (3269, 2541, 1601, 161, and 286 images in Samples 1-5, respectively). Two experts selected 25 uniformly distributed slices throughout each stack and manually marked the boundaries of both sample and lipid pools from each sample. This manual segmentation is performed based on the histologically validated characteristics of the sample and lipid pools visible in micro-CT images (29). The model was trained and validated on just 13 manually marked slices. The remaining manually marked slices were used to evaluate performance. At least one test slice was positioned between every two consecutive training/validation slices. The impact of non-uniform distributions of manual markings and their total number on model performance is discussed in the SI Appendix.

Fig. 3 shows some representative test slices from each sample, the corresponding manual markings (denoted by yellow borders), and model segmentations (marked by red borders). Performance was evaluated based on the Dice and Jaccard similarity coefficients (DSC and JSC, respectively) for sample and lipid pool data, (Fig. 3, Fig. 4), Table 1. Samples 1-4 had mean DSC and JSC scores above 0.85 and 0.75, respectively for the lipid pool segmentation and above 0.95 and 0.91, respectively, for sample segmentations, (Fig. 4 A-D). In contrast, the fifth sample exhibited 5 and 7% reduction in the mean DSC and JSC values for the lipid pool segmentation and 5 and 10% reduction in the mean DSC and JSC values for sample segmentation, respectively. The substantially smaller cross-sectional areas of the fifth sample along with the relatively complex shape, (Fig. 3 Q1, R1, S1, T1), caused the segmentation scores to be highly sensitive to even small mismatches between the segmentation maps and the manual markings, resulting in this reduction in segmentation accuracy (30).

Average training times for each sample were less than 45 minutes (Table 1), with significantly more time spent on the sample compared with the embedded lipid pool (Expert 1: lipid pool mean of 331s versus sample mean 2291s, $P = 0.009$; two-sided Wilcoxon rank sum test; Expert 2 lipid pool mean of 333s versus sample: mean 2309 s, $P = 0.009$; two-sided Wilcoxon rank sum test). The average training time for the two experts was not significantly different (Expert 1: mean of 2623s, Expert 2: 2642s $P = 0.754$; two-sided Wilcoxon rank sum test).

### 2.2.1 Technical challenges for segmenting µ-CT data with calcification and embedded lipid pools.

In general, µ-CT images of calcified vascular tissues have several categories of imaging artifacts (31). We briefly review these artifacts and how the framework was designed to address these challenges.

**Low border contrast between lipid pool and surrounding tissue:** While the lipid pool regions are darker than the surrounding sample, the interface between the lipid pool and surrounding tissue are not distinct, (Fig. 3 Sample 1: A1, C1; Sample 2: E1; Sample 5: Q1- T1), hindering accurate segmentation of this boundary in some images. The lipid pool occupies a 3D volume with gradual morphological changes across the sequential 2D µ-CT images (not shown). The shape continuity is leveraged through inclusion of the z position as a feature in the neural network classifier, enabling linking information in neighboring images with sharper boundaries to improve segmentation in those with hazy boundaries, (Fig. 3 A3, C3, E3, Q3- T3), yielding average DSC scores above 0.8 for the lipid pools and 0.92 for the sample boundary (Table 1 – Expert 1).

**Overlap of macrocalcifications and lipid pools:** Overlap of macrocalcification with the lipid pool borders challenges the automated identification of the lipid pool boundary in some images, (Fig. 3 J1-P1). The continuity of the lipid pool across the sample was again leveraged to address this issue and enabled accurate segmentation of the boundary, (e.g., Fig. 3 J3- P3).

**Streak artifacts:** Large calcifications, due to their denser material compared to the adjacent tissues, can induce intense streak artifacts creating dark regions around the calcifications such as those identified with red arrows in Sample 3, (Fig. 3 K1, L1), as well as dark/bright ray-like artifacts such as those in Sample 4, (Fig. 3 M1- P1). Although these artifacts could severely impact the visibility of tissue or lipid pools in the affected regions in some cases, the proposed framework was able to effectively identify the sample and lipid pool regions, (Fig. 3 K2- P2, and K3-P3, respectively), Table 1.

**Ring artifacts:** The most common artifacts present in μ-CT images that can significantly reduce their quality are ring artifacts (31). Their presence hinders accurate segmentation of boundaries of ROIs. Samples 3 and 4 are severely affected by ring artifacts (center marked by cyan arrow), (Fig. 3 K1, L1) and (Fig. 3 M1-P1), respectively. Despite the overlap of ring artifacts and both tissue and lipid pool regions, the segmentation results for the sample and lipid pool boundaries are in good agreement with the manual markings, (Fig. 3 K2-P2) and (Fig. 3 K3-P3), respectively. This demonstrates the capability of the proposed framework for handling such cases without the need for performing any ring artifact reduction techniques.

**Undesired presence of sample holder:** Sample holders are required to constrain samples inside sample container during scans using μ-CT. In some cases, these holders may appear in μ-CT images at a grey scale similar to that of the sample, green arrows in (Fig. 3 A1-E1, Q1), hindering segmentation of ROIs using the image thresholding approach. Our segmentation algorithm effectively leverages the spatial coordinates associated with each pixel so that pixels outside the ROI can be removed, e.g. (Fig. 3 A2-E2, Q2).

## 2.3 Calcification phenotyping based on size, topology and clustering using an ML-based clustering algorithm

A novel framework is developed to classify calcifications based on size, spatial distribution and topology, Fig. 5. First, each calcification particle in the domain is segmented using image thresholding and the volume is calculated based on a prescribed volume threshold after performing 3D reconstructions. In prior work, microcalcifications were defined as having an equivalent diameter of less than 500 um based on the imaging resolution of most clinical scanners, (e.g. Gade, *et al.* (3)). Here, we use this same critical diameter and prescribe all larger particles as macrocalcifications. The microcalcifications are then further subcategorized based on spatial distribution as either isolated or part of a cluster. Clusters are detected through the application of a machine-learning-based clustering algorithm. The remaining microcalcifications are designated as isolated calcifications. In the next step, macrocalcifications are further categorized as sparse or dense based on their topology (Methods).

As shown in Fig. 5, the first specimen had two macrocalcifications, each with sparse and dense parts, and seven clusters of microcalcifications. The second specimen showed no macrocalcifications but had three clusters of microcalcifications. The third sample exhibited two macrocalcifications, each with sparse and dense parts, along with two clusters of microcalcifications. In the fourth sample, one macrocalcification, with both dense and sparse parts, was observed, and this sample did not have any clusters of microcalcifications. The fifth sample displayed one entirely sparse macrocalcification and three clusters of microcalcifications.

## 2.4 Application of the proposed pipeline for high throughput phenotyping of vascular calcification

The two frameworks introduced in Sections 2.1 and 2.3 were integrated into a single pipeline (Fig. 6) to enable phenotyping of samples based on i) co-localization with lipid pools (atherosclerotic/nonatherosclerotic), ii) size (micro/macro), iii) distribution of microcalcifications, and iv) topology of macrocalcifications. The pipeline was applied to the five specimens to classify every calcification particle within the sample by phenotype. Briefly, outputs from the first and second framework were merged to create 3D volumes in which each calcification is identified as within or external to a lipid pool (*atherosclerotic/non-atherosclerotic*). Data on volume of each calcification enable classification by size (*micro/macro*). Microcalcifications were further subdivided as part of a cluster or as isolated. Macrocalcifications were subcategorized by their topology (*dense/sparse*).

The pipeline described in this section was successfully applied to five specimens, enabling valuable insights into the composition and characteristics of atherosclerotic plaques. Fig. 7 provides a visualization of representative outcomes. High fidelity 3D models of tissues, lipid pools, and calcification display the relative position of these components and can be used for further biomechanical studies (Fig. 7A). The relative location of these components can be easily visualized using computer generated cross sections, (Fig. 7 B). The calcification phenotypes across each specimen can be quantified regionally or for each sample, providing essential data for studying the relationship between pathology and rupture risk, (Fig. 7C) and further broken down for macro (Fig. 7D) and microcalcifications (Fig. 7E).

In Fig. 7 C, we present the volumetric ratios of lipid pools to tissue and calcifications to tissue, indicating the extent of atherosclerosis and calcification, respectively. Additionally, we provide the volumetric ratio of atherosclerotic calcifications to the entire calcification domain, a crucial factor for rupture risk assessment (3). This ratio is particularly important, as previous studies have suggested tissues with atherosclerotic calcifications tend to exhibit increased stability compared to those with non-atherosclerotic calcifications (3). Furthermore, it is known that lipid pools can attenuate stress concentrations around calcifications (16), further emphasizing their relevance in the context of plaque stability (Fig. 7 C). The proportion of macrocalcifications to the entire calcification domain is also presented in Fig. 7 C, highlighting another essential risk factor for plaque failure. Previous research has indicated macrocalcifications can stabilize plaques by reducing adjacent tissue deformability, while microcalcifications may induce aberrant stresses that could lead to rupture (4).

Furthermore, Fig. 7 C presents the volumetric ratio of clusters of microcalcifications to the entire calcification domain. This measure is crucial for understanding the amplified deleterious impact of densely distributed microcalcifications on the strength of diseased tissues (6).

In Fig. 7 D, we explore the proportion of sparse and dense parts within atherosclerotic and non-atherosclerotic macrocalcifications. This information is particularly valuable for studying whether the presence of calcifications inside lipid pools is inversely correlated with their topology (sparsity). Higher density (less sparsity) is associated with smoother surfaces and fewer sharp edges, while sparse macrocalcifications due to their sharp edges may create regions with high-stress concentrations and promote failure (16). Investigating the association between the presence of lipid pools and macrocalcification topology is crucial for understanding plaque stability (Fig. 7.D).

Lastly, Fig. 7 E examines the proportion of isolated and clustered microcalcifications within atherosclerotic and non-atherosclerotic macrocalcifications. This data is essential for examining correlations between the presence of lipid pools and the distribution density of microcalcifications. It also addresses the question of whether the presence of lipid pools facilitates the agglomeration of microcalcifications and accelerates micro-to-macrocalcification conversions (15). Additionally, while higher distribution density of microcalcifications is linked to amplified deleterious impacts on tissue stability (6), the presence of these microcalcification clusters inside lipid pools may reduce concerns about their role in the failure process, given that lipid pools can attenuate stress concentrations around these particles. Understanding the role of lipid pools in the presence of microcalcification clusters is crucial for refining rupture risk assessments (Fig. 7 E).

The time durations required to complete each step of the pipeline after obtaining the µ-CT images for each specimen (Fig. 6 D) are presented in Fig. 7 F. The image segmentation time was calculated by combining the required training and application time of the proposed segmentation framework for sample and lipid pool segmentation with the time needed to perform calcification segmentation via image thresholding. The required time for segmentations using the proposed framework was presented in Table 1. Additionally, Fig. 7 F displays the time required for calcification segmentation. It is evident from this figure that sample segmentation is the most time-consuming component of the overall segmentation time, whereas calcification segmentation time is significantly shorter than that of sample or lipid pool segmentations due to the use of image thresholding. Furthermore, for samples with larger macrocalcifications such as the first and the third samples a significant portion of the overall runtime is allocated to particle identification and the classification of macrocalcifications based on their topology (sparse/dense). The runtime required to perform these analyses can be substantially improved by vectorizing the developed algorithms to harness the power of GPU parallel processing. This project is currently underway and will be incorporated into future research endeavors to expedite the analysis.

**2.5 Extending the clustering algorithm to analyze the coupling between collagen fibers and microcalcification density**

Collagen fibers are the central load bearing component in soft biological tissues such as arteries. However, the integrity of the fibers cannot be imaged in vivo and therefore, it is important to identify surrogate markers for collagen integrity that are measurable in patients. A recent in vivo experimental study has demonstrated a tendency for microcalcifications to coalesce within areas of fibrous tissue lacking a dense collagen fiber distribution (15). However, it is not yet well-understood whether the dense distribution of microcalcifications degrades collagen fiber network (causative), or these micro particles amalgamate only in regions with diminished collagen fiber distribution density. In any case, there should be an inverse correlation between the distribution density of microcalcifications and that of collagen fibers. Identifying such correlations requires quantifying the distribution density of both collagen fibers and microcalcifications.

Here, we extend the prior clustering algorithm to collagen fibers to explore the relationship between density of collagen fibers and calcification. In particular, we identify a distribution density threshold for microcalcifications above which collagen fiber content is diminished (Fig. 8). A multiphoton microscopy (MPM) dataset for Sample 1 is used for this example, as this modality can image both collagen fibers and calcification (3), (Fig. 8 A). In the first step, each of the collagen and calcification channels were extracted (Fig. 8 B and 8 C, respectively) from the input dataset and their 3D models were reconstructed by applying 3D triangulation algorithm (Fig. 8 D and 8 E, respectively).

The 3D model of collagen channel is first divided into regions of low and high fiber density using the machine-learning-based algorithm, (Fig. 8 D). Next, the 3D model of calcifications is superimposed on that of the collagen channel, (Fig. 8 E), and calcifications are labeled C1 and C2, based on their co-location with the low and high-density collagen domains, respectively. Qualitatively, regions of low collagen density appear to coincide with regions of high calcification density. The ML-based clustering algorithm is then applied to the calcifications to cluster them into regions of low and high density. The calcifications within these regions are labeled as D1 and D2, respectively, (Fig. 8 F). The ML-based clustering algorithm parameters are iteratively adjusted to obtain an accuracy of 80% or more for the match between D1-C1 and D2-C2 particles, collectively, providing a threshold for calcification density corresponding to diminished collagen content.

**3. Discussion**

There is a strong association between the presence of arterial calcification and major adverse cardiovascular events (MACE) such as stroke and heart attack as well as to other cardiovascular diseases such as intra and extracranial aneurysms, and aortic valve diseases (2, 3, 9, 19, 32-40). However, substantial controversy exists over the mechanistic role of calcification in these diseases and how to effectively integrate clinical measurements of calcification into risk assessment for these diseases. For example, the calcium score, commonly used in clinical practice, assumes increased total calcification corresponds to an elevated risk of adverse events. However, even this foundational assumption is inconsistent with important findings such as evidence that large calcifications (macrocalcifications) can be protective (8, 11, 12). Moreover, statins are associated with increased calcification, yet are known to lower risk of MACE (21, 22, 41), Research to understand these contradictions points to the need to account for the diverse presentations of calcification, including factors such as size, spatial distribution, topology and nature of the surrounding tissue (4-6, 15, 16, 23).

Therefore, in this work we introduced a new system for categorizing calcification phenotypes. Motivated by studies that provide either direct evidence (mechanistic) (6, 8) or indirect evidence (associative) (3) of the roles of calcification in adverse cardiovascular events, we selected four calcification categories for a total of eight distinct phenotypes. The first category distinguishes calcification by size into micro and macrocalcifications. Concerning macrocalcifications, computational studies suggest they shield arterial plaques by reducing their deformability (4, 8, 12). Conversely, comprehensive studies have consistently provided compelling evidence showing microcalcifications can promote tissue failure (6, 13, 14). In a computational simulation, the presence of even a single spherical microcalcification within the plaque fibrous cap was shown to amplify plaque stresses by a factor of 2.5 (5). Another computational analysis demonstrated that the existence of two microcalcifications positioned in close proximity, regardless of their location within the arterial wall, can significantly elevate intramural stresses, driving the tissue toward failure (6). Therefore, calcifications size is an important factor determining their role in the failure process.

The second and the third categories focus on macrocalcification topology (sparse or dense), and microcalcification spatial distribution (isolated or clustered), respectively. Regarding macrocalcification topology, a recent in silico simulation showed the sharp edges of macrocalcifications can lead to abnormal stress concentrations in surrounding tissues, thereby compromising tissue stability (16). Regarding microcalcification spatial distribution, an in-vivo experimental study found a direct correlation between the locations of the dense distributions of microcalcifications and regions of tissue with diminished collagen fiber density (15). While it is not yet well known if the presence of dense distributions of microcalcifications degrades collagen networks or if these particles coalesce only in regions with lower density of collagen fibers, such focal changes in the collagen fibers density will certainly culminate in lower tissue strength in these locations as these fibers are the main load-bearing components of soft tissues in physiological loading conditions. Therefore, it is important to consider the impacts of macrocalcification topology and microcalcification spatial distribution as phenotypes that can substantially modulate the tissue failure process.

Prior research also suggests that the detrimental impact of calcifications can be diminished when they are embedded within lipid pools (10, 37). Furthermore, in calcified soft tissues such as aortic valve leaflets and arteries these "atherosclerotic calcifications", i.e., calcifications embedded in lipid pools, are believed to have a different etiology compared with those distinct from lipid pools (42). For this reason, we included a fourth calcification category of atherosclerotic/non-atherosclerotic calcification.

Having identified these categories, we then developed a framework that makes it possible to phenotype thousands of individual calcification particles across vascular specimens. This objective was achieved using two machine-learning-based frameworks. The first framework leverages deep learning approaches to perform the segmentation of tissue components. The second framework is designed to classify calcifications based on their size, spatial distribution and topology. Together, these two frameworks provide a powerful and automated means of distinguishing all 8 phenotypes in diseased tissues, which can have significant implications for the study and diagnosis of calcification-related diseases.

Regarding the first framework, the segmentation of lipid pool regions in µ-CT images of calcified vascular tissues can be challenging due to their lower contrast and blurred boundaries compared to other regions. Traditional histological analysis is labor-intensive and destructive. Our high-throughput pipeline offers a more efficient, and accurate analysis of intact specimens, improving calcification phenotype quantification. While a research letter published in 2021 also endeavored to present a fully automated segmentation methodology for dissected atherosclerotic tissue based on UNet, the segmentation accuracy for the lipid pool measured by the dice similarity coefficient was 0.41 (43). Furthermore, the impact of CT artifacts and the issue of overlapping macrocalcifications and lipid pools were not discussed, and the segmentation maps were not presented. The low segmentation score for the lipid pool, despite the using 1601 training slices, highlights the importance of employing a semi-automated segmentation approach for accurate segmentation of tissue components in these µ-CT images.

Our frameworks enable more precise investigations into the effects of different treatments on the development of calcifications and lipid pools within vascular tissues, such as those aimed at reversing the calcification process in artery walls (24). It is crucial to acknowledge that not all calcification phenotypes are harmful, and eliminating all calcifications may not necessarily improve tissue stability. For example, if removal of macrocalcifications in arterial stenosis through treatments outpace the natural tissue remodeling process, a void or area incapable of bearing normal loading conditions can be created where macrocalcifications once existed. This situation can transfer excessive stress onto fibrous caps and contribute to their failure. Therefore, it is essential to measure the pace of calcification elimination by these treatments and quantify their impact on intramural stresses associated with different calcification phenotypes.

The pipeline supports both computational and experimental analyses aimed at understanding the complex interplay of various calcification phenotypes on intramural stress within vascular tissues. Such information is needed to improve risk assessment for MACE and other diseases as well as to assess the potential of novel therapies. For computational studies, the pipeline provides detailed 3D subject-specific tissue, lipid pool and calcification models to create necessary computational domains. In experimental analyses, it facilitates selection of regions of interest for mechanical testing and enables the quantification of dissected vascular tissue samples based on calcification phenotypes. This assessment can serve two primary purposes: first, to investigate the relationship between calcification and tissue strength through mechanical testing, shedding light on the role of calcification in tissue failure; and second, to evaluate the efficacy of innovative therapeutic interventions designed to target vascular calcifications (24).

Here, we focused on eight phenotypes to demonstrate the approach and methodology. This approach can easily be adapted to increase or even decrease these categories as scientific studies proceed and more direct and indirect evidence is available as to the role of calcification in MACE. For example, here we utilize two size categories using a threshold for microcalcifications motivated by the resolution of clinical scanner (44). Additional subcategories can be introduced and the threshold size for microcalcifications can be modified depending on the application and future findings. Previous studies have reported a correlation between the presence of micro and macrocalcification and location within the vascular wall (3). While we have not introduced a category for the physical location of calcifications across the wall thickness, we have considered this in a prior study (45) and it can be useful for future investigations (3).

Currently, standard clinical CTA scanners cannot detect microcalcifications due to resolution limitations on the order of 0.5 mm (44). Therefore, tools such as the calcium scoring (CAC) with the current clinically implemented measurement techniques, cannot incorporate the impact of the microcalcifications and their phenotypes such as clustered or isolated (based on their spatial distributions). This may constitute one of the reasons for CAC inconsistencies such as the high false positive rate (46) and diminished accuracy for younger patients (20). Among the calcified vascular tissues considered in this study (Fig. 7 C), sample 3 had a greater calcification extent compared to sample 1, which is evidenced by their calcification to tissue volumetric ratios (11.5% vs 2.1%), while sample 1 had substantially more microcalcifications than sample 3 (5509 vs 104). Furthermore, as illustrated in Fig. 7 E, sample 1 had more non-atherosclerotic clusters of microcalcifications than sample 3 (17.8% vs ~0%). Evaluating the vulnerability of these samples using the current calcium scoring, these samples will be stratified equally as high-risk, due to its sensitivity to only macrocalcifications and ignoring the impact of microcalcification. However, based on the most-recent findings, sample 3 has a significantly lower risk of failure (6, 12, 16).

Recently, new intravascular imaging modalities have been introduced with the capability of imaging the inner wall of larger arteries (27). One such modality, intravascular micro-optical computed tomography (µ-IVOCT) enables in-vivo detection of microcalcifications and therefore has the potential to provide the necessary data to improve the calcium scoring index. In particular, our high throughput quantification tool could be implemented for these datasets and provide the corresponding calcification phenotypes along the vasculature. Such phenotyping could be used to enhance the diagnostic value of risk assessment tools once a mechanistic understanding of the role of the calcification phenotypes has been determined. While we have motivated the present work by applications to MACE, calcification is also found in heart valves, peripheral vessels and many other soft tissues across the circulatory system (1, 43, 47). The approach introduced here is equally applicable to calcification in other areas and can be tailored to the relevant phenotypes for a particular disease.

In summary, the presented classification system for vascular calcifications and the novel quantification tool enable analysis of calcification in diseased tissues and represent an important step towards improving risk assessment tools for MACE and other calcification-related diseases. Specifically, the proposed pipeline quantifies specimens based on type, size, topology, and spatial distribution of calcifications. Our AI tool simplifies information extraction for objective analysis of numerous specimens, aiding studies exploring the relationships between calcification phenotypes and disease progression and treatment. High-throughput quantification combined with mechanical testing can be used to identify the role of each calcification phenotype in the failure process. Applying these insights to in-vivo IVOCT imaging (48) will enhance the diagnostic value of risk assessment tools such as calcium scoring, potentially reducing the economic burden of cardiovascular disease.

## 4- Methods
### 4.1- The volumetric µ-CT image segmentation framework

The neural networks employed in this study consisted of a single hidden layer with 500 neurons. For training, we utilized the pixel-wise cross-entropy loss function, as defined in Eq. 1.

$$\text{Cross entropy loss} = - \sum_{i=1}^{N} \sum_{j=1}^{K} t_{ij} \ln y_{ij} \quad \text{(Eq.1)}$$

where *N* is the total number of pixels and *K* is the number of classes, $t_{ij}$ is determined by the ground truth and is a binary indicator of whether pixel i belongs to class j, and $y_{ij}$ is the model-predicted probability for pixel *i* belonging to class j. The cross-entropy loss outperforms other loss functions, such as the dice loss, in cases where the size (surface area) of the foreground experiences significant variations compared to the background (49), as is typically observed in µ-CT stacks of dissected atherosclerotic vascular specimens. For the hidden layers and output layer, we employed the ReLU and SoftMax activation functions, respectively. To prevent overfitting, we implemented an early stopping convergence criterion based on validation accuracy, halting the training process if there was no improvement in validation accuracy over 45 and 15 consecutive epochs for sample and lipid pool segmentations, respectively. The limited-memory Broyden-Fletcher-Goldfarb-Shanno quasi-Newton algorithm (LBFGS) (50) was employed to minimize the cross-entropy loss function for both networks.

The segmentation of calcifications is carried out by image thresholding approach. A global threshold was applied to the color intensity of pixels in the input grayscale images, with all pixels exhibiting a color intensity greater than this threshold being labeled as calcifications.

### 4.2- Evaluation Metrics

We considered the Dice similarity coefficient (DSC) and Jaccard similarity coefficients to evaluate the sample and lipid pool segmentation accuracy, which is defined as:

$$\text{DSC} = \frac{2 \times \text{TP}}{2 \times \text{TP} + \text{FP} + \text{FN}} \quad \text{(Eq.2)}$$

The value of a DSC ranges from 0, indicating no spatial overlap between two sets of binary segmentation results, to 1, indicating complete overlap.

$$JSC = \frac{TP}{TP+FP+FN} \tag{Eq.3}$$

The range of JSC span values between 0 and 1, in which JSC = 0 means there is no overlap between the segmentation map and the ground truth and the value of 1 indicates perfect match.

### 4.4- Quantification and Statistical Analysis.

The statistical analysis of the model performance comparisons in this study was conducted using either the two-sample t-test or two-sided Wilcoxon rank sum test depending on the number of measurements. Unless otherwise stated, results were considered statistically significant only if the calculated P values were less than 0.05.

### 3.5- Data and software availability

The entire proposed pipeline is developed in MATLAB programming language. The proposed semantic segmentation framework in this study is also developed in Python using both TensorFlow and PyTorch libraries. The full code used in the experiments isn't publicly accessible at this time, but the network design details are included in the results and Materials and Methods sections. The source code can be obtained upon request, with approval from the University of Pittsburgh.

### Acknowledgments

We express our gratitude to the University of Pittsburgh's Brain Bank for generously providing the cerebral arteries and aneurysms utilized in this study. Additionally, we extend our appreciation to the National Institutes of Health for their invaluable support through NIH grants (2R01NS097457, 1S10OD025041), which significantly contributed to the execution of this research. We are indebted to the invaluable contributions of Reza Ramezanpour (Department of Mechanical Engineering, Purdue University), Fatemeh Azari (Department of Mechanical Engineering and Materials Science, University of Pittsburgh), as well as Gabrielle Wassel, Patrick Tatlonghari, and Apurva Rane (Department of Bioengineering, University of Pittsburgh) for their meticulous manual annotations of the micro-CT datasets.

**Figures and Tables**

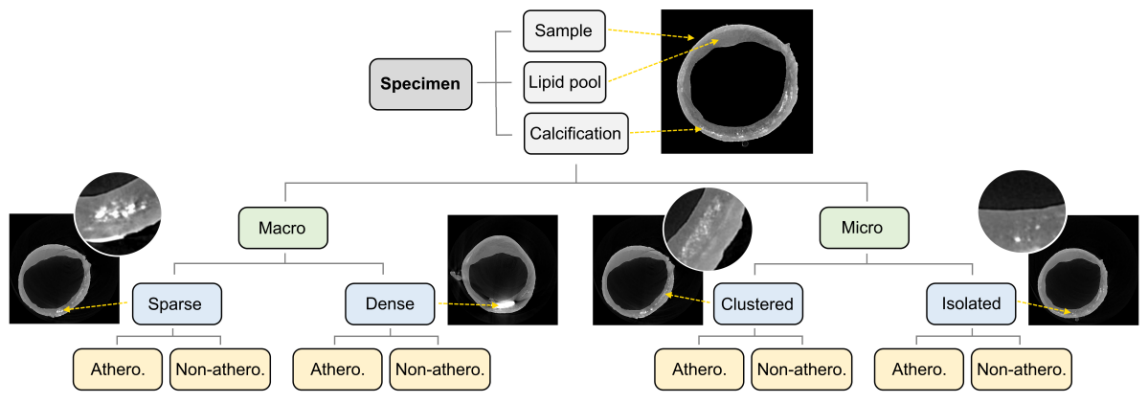

**Figure 1.** A classification system for phenotyping calcifications in human vascular tissues. This process begins by categorizing calcifications into two groups based on their size: microcalcifications and macrocalcifications. Macrocalcifications are further examined for their topology and sorted into either sparse or dense categories. The next step involves evaluating microcalcifications based on their spatial distribution, utilizing an unsupervised clustering algorithm to differentiate between clustered and isolated occurrences. Lastly, these categories are all further sub-categorized based on their colocation with lipid pools as athero- or non-atherosclerotic leading to eight distinct phenotypes.

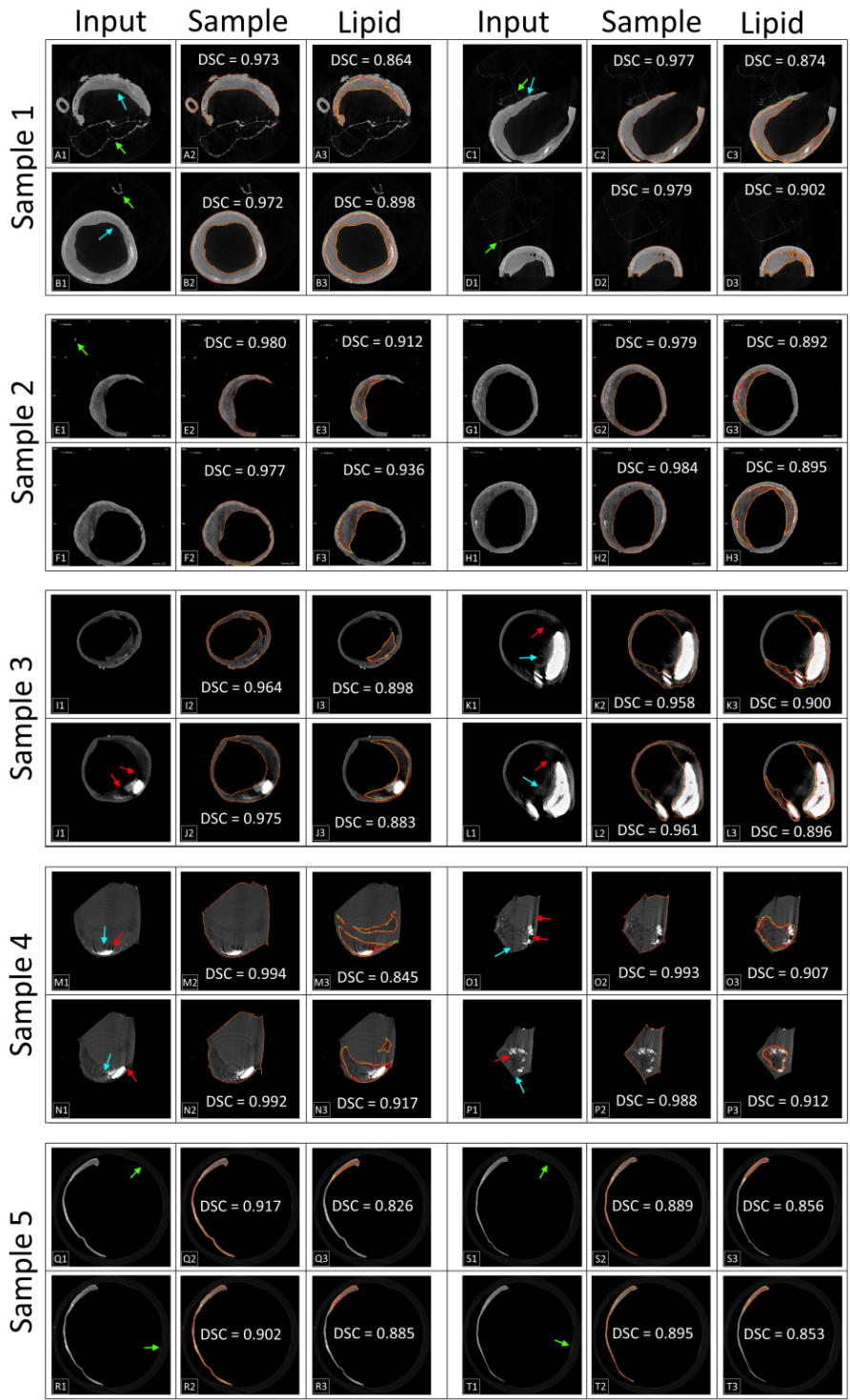

**Figure 3.** Comparison of the model segmentations with manual markings in representative cross sections from the five samples. Column 1 and 4: raw input µ-CT slices, Column 2, 5, and 3, 6: manual markings (yellow) and model segmentation output (red) for the sample and lipid pool boundaries, respectively. Bright white regions are calcification. Artifacts include streak artifacts due to large calcifications (red arrows), ring artifacts (cyan arrow at ring center), undesired presence of sample holder (green arrows). Ring and streak artifacts can significantly hinder accurate segmentation, particularly for lipid pools. The algorithm performance was robust despite the significant presence of these artifacts.

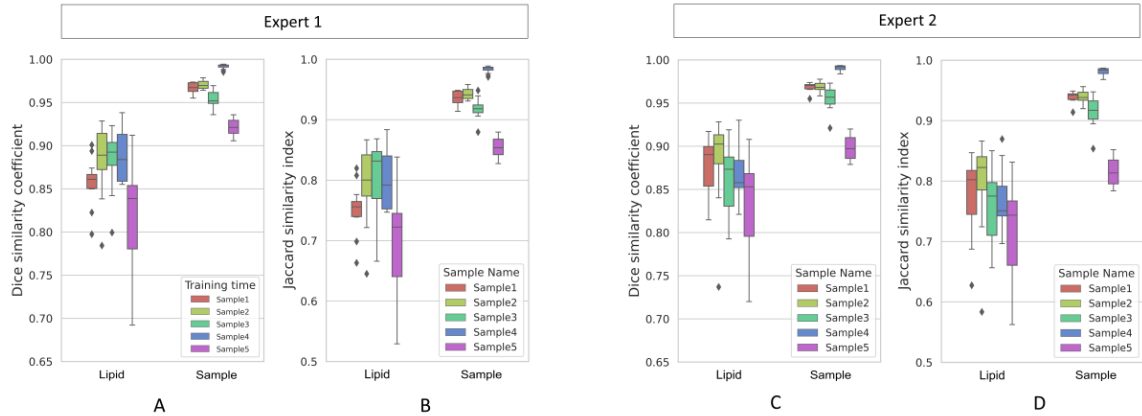

**Figure 4.** Overall performance of the proposed segmentation framework for sample and lipid pool regions measured by the Dice and Jaccard similarity coefficients for Expert 1 (A and B, respectively) and Expert 2 (C and D, respectively). Experts manually marked 25 uniformly distributed slices in each stack, of which 13 uniformly distributed slices were dedicated to training and validation. The remaining slices were used to test the model performance. Mean DSC and JSC values for the four samples (1-4) were over 0.85 and 0.75 for lipid pool segmentation and over 0.95 and 0.91 in sample segmentation. The fifth sample had a very small cross-sectional area, making the performance metrics highly sensitive to slight non-overlapped pixels between the segmentation and the manual markings. As a result, mean DSC and JSC scores were 0.81 and 0.69 for lipid pool and 0.90 and 0.81 for sample segmentation in this sample, respectively, Table 1.

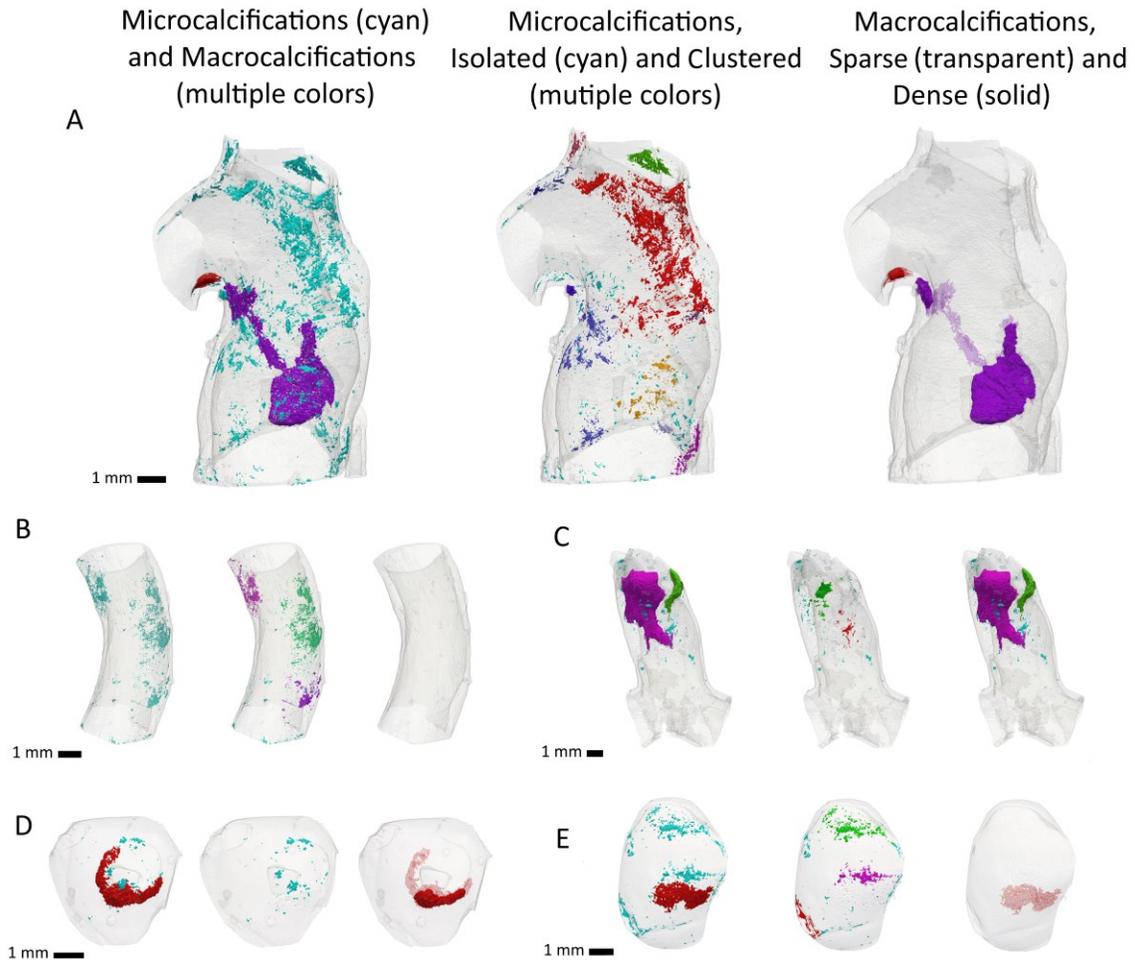

**Figure 5.** Calcification classification based on size and distribution density using the proposed framework for the five samples. The input calcification domains are analyzed to identify microcalcifications (cyan) and macrocalcifications (multiple colors except cyan), (column 1). The microcalcifications are inspected for the existence of the clusters of microcalcifications (column 2, distinct colors for each cluster, isolated particles in cyan), and each macrocalcification is then further processed to assess topology (column 3, sparse parts in transparent colors, and non-transparent colors for the dense parts).

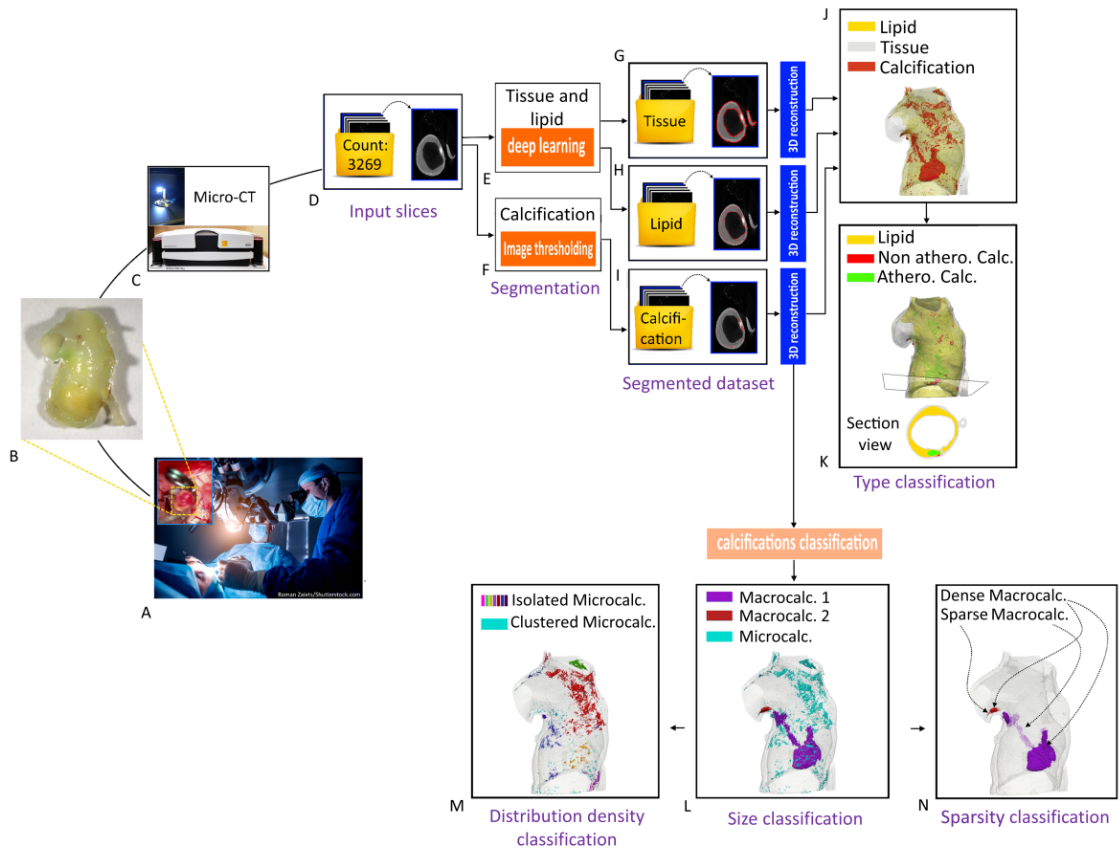

**Figure 6.** Pipeline for quantification of calcification based on size, spatial distribution of microcalcifications, topology of macrocalcifications and co-localization with lipid pools. (A) Vascular specimens are harvested from surgical procedures or from cadavers. (B) Representative cadaveric cerebral artery sample. (C) Samples are scanned via high-resolution µ-CT imaging to generate (D) stacks of reconstructed grey scale images. (E,G,H) Sample and lipid pool regions are semi-automatically identified across the entire stack via the segmentation framework. (F) Calcifications are segmented using image thresholding. (J) 3D reconstruction of the tissue, lipid pools and calcifications are performed for each sample using the 3D triangulation algorithm. (K) Atherosclerotic and non-atherosclerotic calcifications are identified based on their co-localization with lipid pools. (L) All classifications are categorized by size as micro or macrocalcifications. (M) Microcalcifications are further categorized as isolated or as part of cluster. (N) The volume of each of the identified macrocalcifications are further classified as sparse or dense.

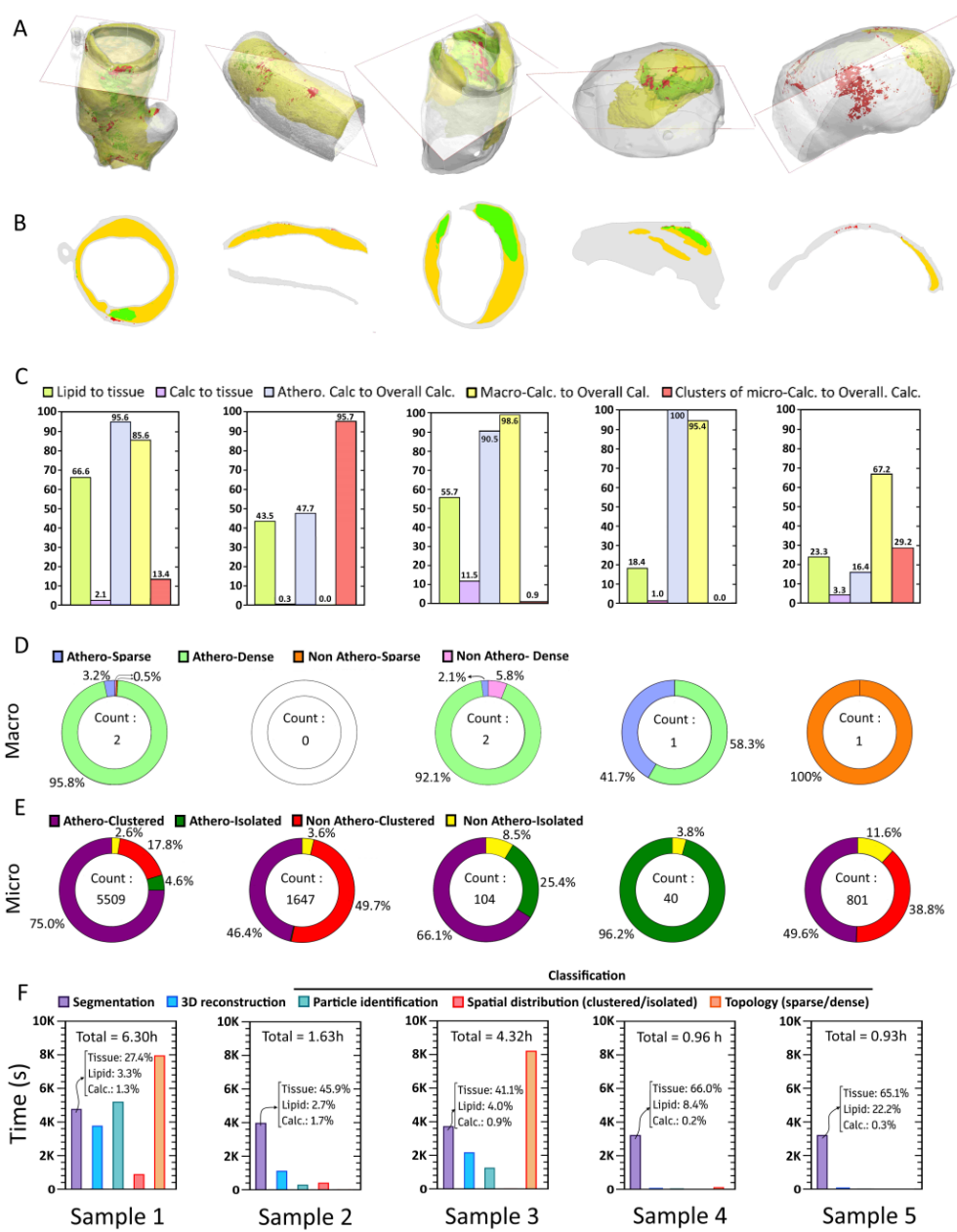

**Figure 7.** The outcomes of deploying the proposed pipeline to quantify the calcified vascular specimens considered in this study. (A) the 3D reconstructions of the sample (white color), lipid pools (yellow regions), and calcifications (red for non-atherosclerotic calcifications and green particles for atherosclerotic ones). (B) the cross sections of the 3D models using the planes shown in the first row are displayed, showing the location of lipid pools, atherosclerotic and non-atherosclerotic calcifications with the tissue walls. (C-E) Representative compiled data resulting from applying pipeline to five samples. (F) The required time to complete each step of the proposed pipeline (after acquiring high-resolution µ-CT images) to perform specimens quantifications. The segmentation time comprises the time required to train the proposed semi-automatic segmentation framework to segment sample plus lipid pool regions, application, and performing calcification segmentation. This panel also presents the proportion of time allocated to training the segmentation model for sample, lipid pool, and calcification within the overall segmentation time. Particle identification presents the required time to detect each calcified particle in the input calcification domain obtained by applying 3D reconstructions to the calcification segmentation results. The required time to classify calcifications based on their size was negligible compared to other steps and is not included in this figure.

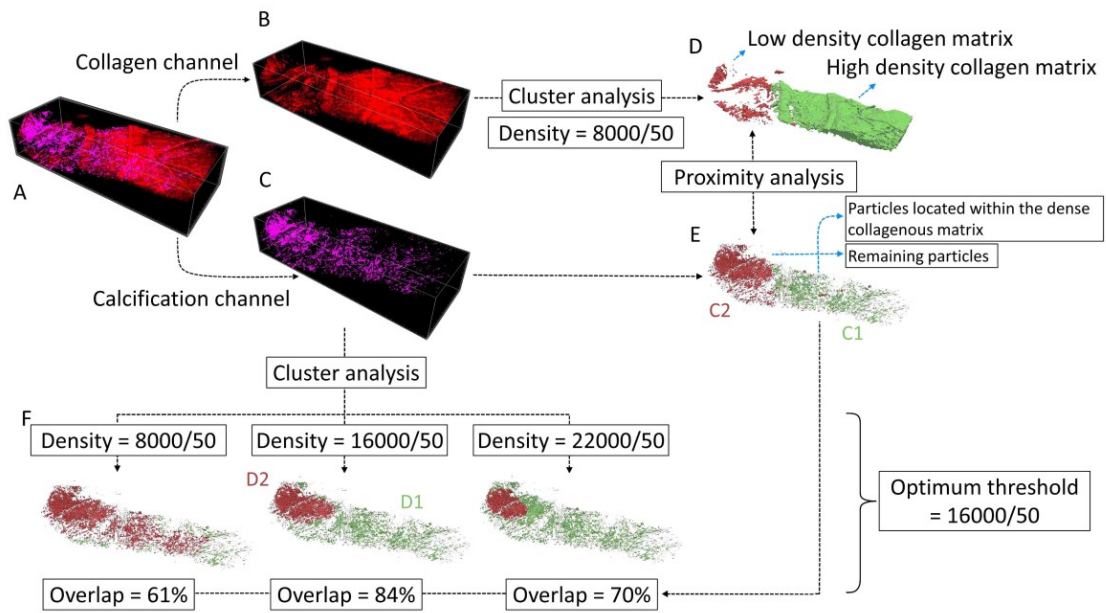

**Figure 8.** Identification of the critical calcification density threshold above which the microcalcifications presence corresponds to regions with diminished collagen fiber structural integrity. (A) Input multiphoton dataset showing the collagen (red) and calcification (magenta) domains. (B and C) The extracted collagen and calcification domains, respectively. (D) Low- and high-density parts of the 3D reconstruction of collagen domain identified using the ML-based clustering algorithm. (E) Identification of calcification particles in domain of low/high collagen density- denoted as C1/C2. (F) Iterative application of the ML-based clustering algorithm to divide calcification into low/high density clusters (D1,D2, respectively). Identification of optimal values of ML-based clustering algorithm parameters to minimize the mismatch between calcification particles labeled C1- D1 and C2 - D2. These parameters provide insights into the correlations between the microcalcification distribution density and locally diminished collagen fiber integrity.

**Table 1.** Summary of the segmentation framework performance for sample and lipid pools

| Sample # | Score (Mean ± STD) | | | | Time (s) | | |
|---|---|---|---|---|---|---|---|
| | Lipid | | Sample | | Overall | Lipid | Sample |
| | DSC | JSC | DSC | JSC | | | |
| colspan Expert 1 | | | | | | | |
| Sample 1 | 0.858 ± 0.026 | 0.752 ± 0.04 | 0.967 ± 0.005 | 0.936 ± 0.010 | 2635.9 | 281.8 | 2354.1 |
| Sample 2 | 0.883 ± 0.040 | 0.793 ± 0.062 | 0.971 ± 0.005 | 0.943 ± 0.009 | 2497.0 | 137.5 | 2359.5 |
| Sample 3 | 0.885 ± 0.034 | 0.805 ± 0.060 | 0.954 ± 0.009 | 0.918 ± 0.017 | 2689.0 | 238.0 | 2451.0 |
| Sample 4 | 0.888 ± 0.028 | 0.800 ± 0.046 | 0.991 ± 0.003 | 0.983 ± 0.006 | 2446.9 | 276.3 | 2170.6 |
| Sample 5 | 0.814 ± 0.068 | 0.692 ± 0.094 | 0.922 ± 0.009 | 0.855 ± 0.016 | 2843.6 | 723.0 | 2120.6 |
| colspan Expert 2 | | | | | | | |
| Sample 1 | 0.878 ± 0.030 | 0.777 ± 0.063 | 0.968 ± 0.005 | 0.938 ± 0.009 | 2919.4 | 346.7 | 2572.7 |
| Sample 2 | 0.884 ± 0.049 | 0.796 ± 0.074 | 0.968 ± 0.006 | 0.939 ± 0.010 | 2304.1 | 136.0 | 2168.1 |
| Sample 3 | 0.862 ± 0.040 | 0.760 ± 0.061 | 0.955 ± 0.013 | 0.914 ± 0.024 | 2605.6 | 200.0 | 2405.6 |
| Sample 4 | 0.867 ± 0.031 | 0.767 ± 0.050 | 0.990 ± 0.003 | 0.980 ± 0.006 | 2674.0 | 536.9 | 2137.1 |
| Sample 5 | 0.834 ± 0.053 | 0.720 ± 0.077 | 0.898 ± 0.014 | 0.816 ± 0.024 | 2708.2 | 445.8 | 2262.4 |

## SI Appendix

**The impact of the number and distribution of training/validation slices on performance scores**

The impact of the number and location of training slices (i.e., expert's manual markings) on sample 3, which is known to be challenging in terms of CT artifacts, is shown in Fig. S1, and the statistical comparisons are presented in Table S1. As illustrated in Fig. S1. A, increasing the number of training slices from 5 to 9 had negligible impacts on segmentation scores in slices 1-18 but resulted in a noticeable improvement in the remaining slices. This discrepancy arises from the significant influence of CT artifacts on the latter portion of the stack, whereas the initial half of the stack experienced a comparatively milder impact from these artifacts. In the absence of such artifacts and any other noises, such as the presence of sample holders, the application of image thresholding can present accurate sample segmentations. In the proposed framework (the sample segmentation part, Fig. 1 C), one of the input features of the neural network classifier is generated by thresholding the input images thus, it will be a strong predictor of the correct labels for sample regions, rendering it straightforward for the framework to gain high segmentation scores in cases void of artifacts by training on a few (in this case, five) training slices. However, accurate sample segmentations in the second half required more training efforts, as evidenced by the substantial improvement in the segmentation scores in this part of the stack by increasing the number of training slices from five to nine.

Table S1 presents the results of the experiment that investigated the impact of the number of training/validation slices on the segmentation scores for sample and lipid pool regions. The results show that increasing the number of training/validation slices from 5 to 9 significantly improved the sample segmentation score ($P = 4E-04$; two-sample t-test). However, further increasing the number of manual markings did not lead to a significant increase in the model's sample segmentation performance ($P = 0.16$, $P = 0.36$, and $P = 0.36$ for increasing the number of manual markings from 9 to 13, from 13 to 16, and from 16 to 19, respectively). This indicates that a mean DSC of 96.6% for sample segmentation for such cases with substantial presence of CT artifacts was achieved by using only nine hand-labeled slices.

Regarding lipid pool segmentation, increasing the number of training/validation slices from 5 to 9 and from 9 to 13 significantly improved the segmentation score ($P = 9.4E-11$ and $P = 0.01$, respectively; two-sample t-test). However, further increasing the number of manual markings did not lead to a significant increase in model performance for lipid pool segmentation ($P = 0.80$ and $P = 0.56$ for increasing the number of manual markings from 13 to 16 and from 16 to 19, respectively). This suggests that a mean DSC of 87.0% for lipid pool segmentation was achieved by using 13 hand-labeled slices.

Comparing the number of hand-labeled slices to converge the performance score for lipid pool and sample segmentations, it is evident that the segmentation of lipid pools is more challenging compared to that of sample regions due to their complex morphological characteristics. This was the main motivation for developing the segmentation framework for the sample and lipid pool segmentation in a serial uncoupled fashion (Fig. 2). As mentioned earlier, in cases where the sample-background borders are not significantly affected by artifacts, sample segmentation can be performed via image thresholding, which is a considerably faster approach than performing this task using a semi-automatic deep-learning-based segmentation algorithm. This is evidenced by the substantially shorter segmentation time of calcifications compared to those of the sample and lipid pool regions shown in Fig. 7 F. This serial architecture provides users with the flexibility to perform sample segmentation with fewer training/validation slices or to replace it with the image thresholding approach, resulting in a substantial reduction in segmentation time and overall pipeline runtime.

Achieving model accuracy beyond the converged value depends on two key factors: (1) The consistency and accuracy of the user's manual markings that were assigned to training and test sets, which can be negatively affected by the CT artifacts, (2) the extent of morphological variations of ROI between two consecutive training slices in the input stack, which is a factor of scan resolution and the specimens length scales.

In cases where increasing the number of training slices does not significantly improve model performance, the first source of error, namely inconsistencies in users' manual markings, is likely to be the dominant factor. These inconsistencies can have varying impacts on segmentation accuracy. When present in the training set, they may challenge the deep-learning model's ability to recognize patterns accurately. Conversely, if present in the test sets, they may reduce the model's prediction accuracy by providing an inaccurate ground truth for comparison. Despite these challenges, our developed segmentation framework achieved satisfactory performance using only 13 training/validation slices on a stack of 1601 highly noised $\mu$-CT images, with mean scores of 96.6% and 87.0% for sample and lipid pool segmentation, respectively.

The second source of error in our segmentation framework arises from substantial morphological variations in regions of interest (ROI) between consecutive training slices. To address this issue, we propose two strategies: increasing the number of training slices uniformly or locally. The effect of increasing the number of training/validation slices uniformly was elaborated previously in Fig. S1 (A and B). The impact of the non-uniform distribution of training slices on the performance of our model is depicted in Fig. S1 C and analyzed statistically in Table S1. We employed a total of 13 training slices for all cases with non-uniform distributions. Our results indicate that the distribution of training slices does not significantly affect the model's ability to segment the sample ($P = 0.42$ for bias towards the beginning of the stack, and $P = 0.11$ for bias towards the end of the stack). Regarding lipid pool segmentation, the non-uniform distribution of training slices biased towards the beginning of the stack did not result in significant changes in segmentation accuracy ($P = 0.67$). However, for the bias towards the end of the stack, deviation from uniform distribution significantly affected lipid pool segmentation accuracy ($P \sim 0.05$). This is mainly due to the greater morphological variations in lipid pools near the beginning of the stack compared to the ending slices. As previously noted, lipid pool segmentation is more challenging than sample segmentation due to their morphological differences. When training slices were biased towards the end of the stack, the segmentation framework had insufficient training data to capture the morphological variations in lipid pool regions near the beginning of the stack, leading to a reduction in segmentation accuracy for these regions.

**Performance evaluation of the proposed framework in contrast to alternative segmentation models**

We compared the performance of our proposed segmentation framework with other commonly used deep-learning models for medical image segmentation, namely, UNet2D and UNet3D, and image thresholding. The results of this comparison are presented in Fig. S2, and the mean segmentation scores and required segmentation times are summarized in Table S2.

Our study demonstrates the superiority of our proposed segmentation framework over other commonly used deep-learning models for image segmentation of dissected vascular specimens, as evidenced by the results presented in Fig. S2 and Table S2. Remarkably, our framework achieved the highest mean scores for both sample and lipid pool segmentation using 16 and 32 training/validation slices. Statistical analysis further confirms the superiority of our framework's segmentation performance in lipid pool and sample segmentations using 16 training/validation slices when compared to other deep-learning models (Fig. S2). However, when 32 training/validation slices were implemented, UNet2D demonstrated comparable performance to our framework in lipid pool segmentations, and our model's performance in sample segmentation was not significantly different from that of UNet2D, UNet3D and image thresholding, as expected.

Our study endeavored to create a robust segmentation framework for a high-throughput specimen quantification tool, guided by three primary objectives. First and foremost, our aim was to establish a segmentation framework that could reliably perform segmentations with minimal manual input from users across all atherosclerotic samples. Secondly, we sought to reduce the time required for segmentation (Fig. 7 F), as this plays a significant role in the overall runtime of the proposed pipeline, and any enhancements in this area could substantially improve the pipeline's speed. Lastly, our objective was to ensure that the application of this framework does not necessitate any prior knowledge beyond the ability to mark lipid pools and sample regions.

With regards to the primary objective, the superior performance of our framework compared to UNet2D at smaller training set sizes can be attributed to the application of transfer learning, which enables the combination of the ability of UNet2D to detect common intra-slice features of regions of interest (ROIs) among training slices and the ability of the fully connected neural network classifier to identify the gradual volumetric development of ROIs throughout the stack of µ-CT images. While UNet2D alone is unable to perceive the volumetric variations of ROIs in the stack of µ-CT images, UNet3D possesses such ability, albeit with a higher computational cost due to the increased dimensionality of the input data. The application of a fully connected network in conjunction with the UNet2D in our segmentation framework allowed linking the correct labels to the pixels' spatial features within the stack of µ-CT images. This is especially helpful in cases where the µ-CT artifacts locally impact the visibility of ROIs in the input stack. In such cases, our proposed framework automatically reduces the weights of features influenced by pixels' color intensities and instead increases the weights of spatial features to detect the ROIs in problematic regions of the stack with diminished visibility by performing interpolations on the adjacent training slices.

Regarding the image thresholding approach, it has been observed that it is not a dependable model due to its inability to effectively differentiate between the sample or lipid pool regions and other undesirable regions that may be introduced by the sample holder or due to CT artifacts. The inadequacy of this technique is further illustrated in Fig. S2 (B and C). In these figures, a comparative analysis is presented between the performance of the proposed framework and the image thresholding approach with regard to sample segmentation in two exemplary slices - one without the presence of any artifacts that may impede the visibility of sample borders (Fig. S2 B) and another with the significant presence of such artifacts (Fig. S2 C). These slices correspond to the maximum (Fig. S2 B) and minimum (Fig. S2 C) segmentation scores attained by the image thresholding approach. As depicted in Fig. S2 B, there exist no noteworthy disparities in the segmentation performance between the proposed framework and the image thresholding approach, notwithstanding the slightly lower segmentation score of the proposed framework in the artifact-free slice. However, in the presence of CT artifacts, the proposed model exhibits a markedly superior segmentation performance as compared to the image thresholding approach. The inaccuracies associated with the image thresholding technique can significantly compromise the precision of 3D reconstructions that are generated from segmentation results. Consequently, despite its speed, the image thresholding approach is deemed unsuitable as a reliable segmentation model for atherosclerotic samples that are frequently afflicted by CT artifacts.

Considering the second objective, in terms of the segmentation time (Table S2), the proposed framework could outperform UNet2D and UNet3D when the number of training/validation slices was increased from 16 to 32 while presenting comparable segmentation accuracies. Using 32 training/validation slices, the proposed segmentation framework's run time was equal to 45.6% and 31.6% of UNet2D's and UNet3D's run time. As reported in Table S2, the proposed framework required more time to perform sample segmentation compared to UNet2D. This is due to combining the UNet2D and a neural network classifier in the proposed framework (Fig. 2 (B and C)). However, for the lipid pool segmentation, the implementation of convolutional layers with predefined weights and excluding the background pixels from the training process of the second neural network classifier (lipid classifier) (Fig. 2 K) via the foreground-pass filter in the proposed framework (Fig. 2 F), resulted in a substantial reduction in the segmentation time. Therefore, the improved run time for lipid pool segmentation via the proposed framework outweighs its extended runtime for sample segmentation and results in its superior overall runtime compared to UNet2D and UNet3D.

Considering the last objective, both the proposed framework and UNet2D provide a straightforward application procedure for users. However, the utilization of UNet3D requires the users to meticulously arrange the 3D blocks of the test slices in a particular sequence, which conforms to the arrangement of the 3D training and validation blocks. This requirement limits the applicability of UNet3D to situations where the training/validation slices are uniformly distributed; otherwise, the complete coverage of the test sets through the 3D blocks might not be feasible. The rationale behind this issue is the incorporation of max-pooling layers in UNet3D's architecture with a stride of 2 in all directions, which can be resolved by reducing the stride size to 1 along the inter-slice direction. However, applying such modification to the UNet3D architecture will inevitably result in a substantial increase in its overall segmentation time.

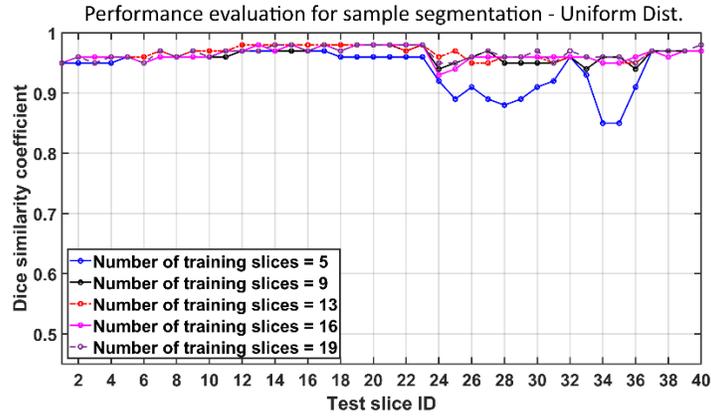

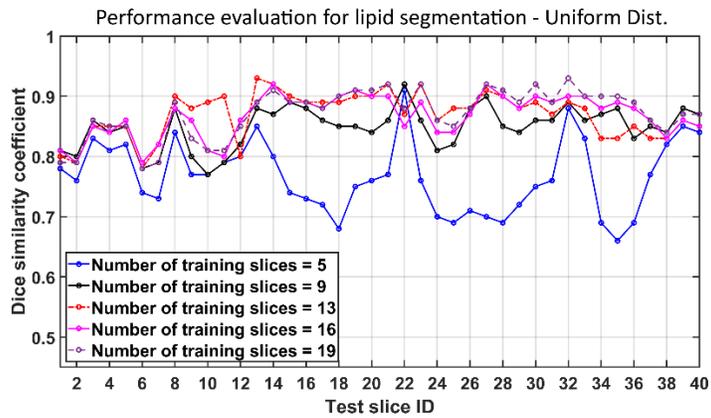

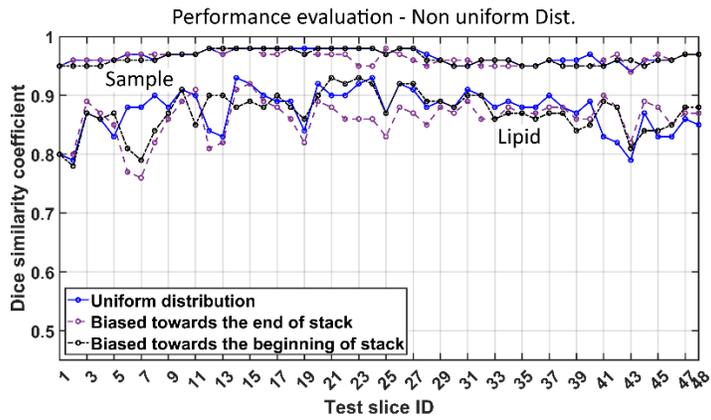

**Fig. S1.** The effect of increasing the number of manually marked training slices on proposed framework performance measured by the dice similarity coefficient. (A and B) The performance score when the number of training slices is increased uniformly from 5 to 19 for the sample and lipid pool segmentation, respectively. (C) The performance score, considering a non-uniform distribution of training slices (the number of training slices in all cases is 13).

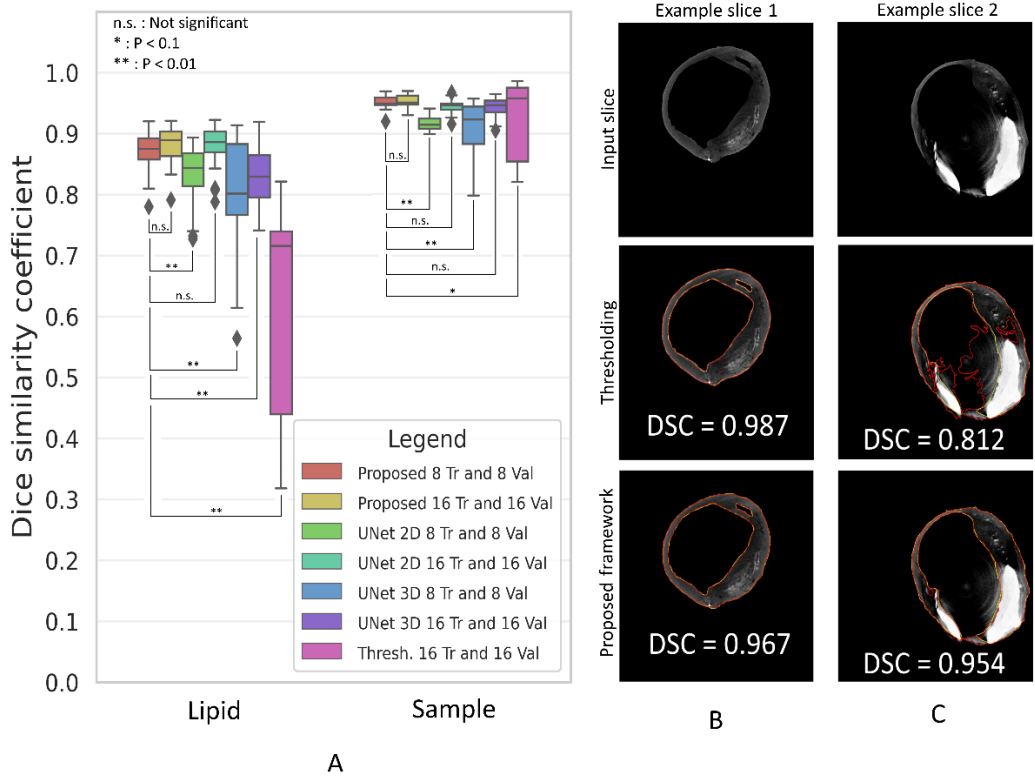

**Fig. S2.** Comparison of the performance of the proposed model with three common segmentation models, including non-volumetric and volumetric UNet and image thresholding approach (A). In terms of lipid pool segmentation, the proposed framework, using 16 training\validation slices, achieved a comparable performance score to the non-volumetric UNet with even 32 training\validation slices. In terms of sample segmentation, using 16 training\validation slices, the proposed framework outperformed other segmentation models at the significance level of 0.1. At the significance level of 0.01, the image thresholding approach had comparable performance to that of the proposed framework. The segmentation maps for two slices corresponding to the maximum and minimum segmentation scores of the thresholding approach are compared to those of the proposed framework in (B) and (C), respectively. As illustrated in these figures, there were no noticeable differences in the segmentation maps for high scores (B). However, for low scores, the segmentation maps were significantly different (C). Such inaccurate segmentations presented by the image thresholding approach, despite having a marginal impact on the mean segmentation score across all slices, hinder precise 3D reconstruction and accurate quantifications.

**Table S1.** Evaluation of the dependence of model performance on the number and distributions of training slices

| | Mean DSC ± SD | | | | | | | |
|---|---|---|---|---|---|---|---|---|
| | Uniform distribution | | | | | Uniform vs. non-uniform distribution | | |
| **Manual markings** | 5 | 9 | 13 | 16 | 19 | 13 (uniform) | 13 (biased to the beginning of stack) | 13 (biased to the end of stack) |
| **Sample** | 0.942 ± 0.034 | 0.962 ± 0.012 (**) | 0.966 ± 0.011 (n.s.) | 0.963 ± 0.012 (n.s.) | 0.966 ± 0.010 (n.s.) | 0.966 ± 0.011 | 0.964 ± 0.012 (n.s.) | 0.963 ± 0.010 (n.s.) |
| **Lipid** | 0.766 ± 0.060 | 0.848 ± 0.035 (**) | 0.869 ± 0.038 (*) | 0.867 ± 0.035 (n.s.) | 0.872 ± 0.041 (n.s.) | 0.875 ± 0.036 | 0.872 ± 0.035 (n.s) | 0.861 ± 0.034 (*) |

(n.s.) Not significant, (*) P < 0.05, (**) P < 0.01. For uniform distribution, the significance levels are compared to the previous column. For non-uniform distributions, comparisons are made with respect to the uniform distribution.

**Table S2.** Comparative analysis of semantic segmentation models: training time and segmentation scores.

| Algorithm | Number of training/validation slices | Score (Mean ± STD) | | Time (s) | | |
|---|---|---|---|---|---|---|
| | | Lipid | Sample | Overall | Lipid | Sample |
| **Proposed framework** | 8 Tr, 8 Val | 0.868 ± 0.034 | 0.951 ± 0.010 | 3067.31 | 236.43 | 2830.88 |
| **Proposed framework** | 16 Tr, 16 Val | 0.879 ± 0.033 | 0.954 ± 0.011 | 5995.94 | 526.96 | 5468.98 |
| **UNet2D** | 8 Tr, 8 Val | 0.828 ± 0.053 | 0.915 ± 0.011 | 4707.04 | 2603.30 | 2103.74 |
| **UNet2D** | 16 Tr, 16 Val | 0.876 ± 0.036 | 0.944 ± 0.011 | 13155.51 | 7905.95 | 5249.56 |
| **UNet3D** | 8 Tr, 8 Val | 0.798 ± 0.098 | 0.913 ± 0.040 | 9287.94 | 6224.73 | 3063.21 |
| **UNet3D** | 16 Tr, 16 Val | 0.831 ± 0.047 | 0.943 ± 0.015 | 18,978.63 | 13156.89 | 5821.74 |
| **Thresholding** | 16 Tr, 16 Val | 0.620 ± 0.169 | 0.925 ± 0.987 | 75.67 | 36.06 | 39.61 |